\documentclass[runningheads]{llncs}
\usepackage{amsmath}
\usepackage{amssymb}
\usepackage{graphicx}
\usepackage{subcaption}
\usepackage{tabularx}
\usepackage{booktabs}
\newcommand{\etal}{\textit{et al.~}}
\newcommand{\ie}{\textit{i.e.~}}
\newcommand{\cd}{ContrastDiagnosis~}

\begin{document}
\title{ContrastDiagnosis: Enhancing Interpretability in Lung Nodule Diagnosis Using Contrastive Learning}
\titlerunning{ContrastDiagnosis}
%
\author{Chenglong Wang\inst{1} \and Yinqiao Yi\inst{1} \and Yida Wang\inst{1} \and Chengxiu Zhang\inst{1} \and Yun Liu\inst{2} \and Kensaku Mori\inst{3,4} \and Mei Yuan\inst{5} \and Guang Yang \inst{1}}
\authorrunning{Chenglong Wang et al.}
%
\institute{Shanghai Key Laboratory of Magnetic Resonance, East China Normal University \and
NVIDIA Corp. \and
Graduate School of Informatics, Nagoya University \and
Center for Medical Big Data Research, National Institute of Informatics \and
The First Affiliated Hospital of Nanjing Medical University
\email{\{clwang,gyang\}@phy.ecnu.edu.cn}}
\maketitle
\begin{abstract}
With the ongoing development of deep learning, an increasing number of AI models have surpassed the performance levels of human clinical practitioners. However, the prevalence of AI diagnostic products in actual clinical practice remains significantly lower than desired. One crucial reason for this gap is the so-called `black box' nature of AI models. Clinicians' distrust of black box models has directly hindered the clinical deployment of AI products. To address this challenge, we propose \textit{\cd}, a straightforward yet effective interpretable diagnosis framework. This framework is designed to introduce inherent transparency and provide extensive post-hoc explainability for deep learning model, making them more suitable for clinical medical diagnosis. \cd incorporates a contrastive learning mechanism to provide a case-based reasoning diagnostic rationale, enhancing the model's transparency and also offers post-hoc interpretability by highlighting similar areas. High diagnostic accuracy was achieved with AUC of 0.977 while maintain a high transparency and explainability.

\end{abstract}
\section{Introduction}

Artificial intelligence (AI), especially deep learning, is changing the way clinicians diagnose diseases. These AI models can sometimes diagnose health problems more accurately than clinicians. Despite this, clinicians still don't use many AI tools in their daily clinical diagnosis. One big reason is that these AI systems are complex and it's hard to see how they come up with their answers, which makes doctors unsure if they can trust them. Clinicians often view these models as `black boxes' that offer little to no insight into the reasoning behind their decisions. This transparency gap raises concerns about accountability and trust, which are fundamental to clinical practice and patient care.

Interpretable AI models are proposed to address these concerns by providing explanations for their predictions that are understandable to human experts. The increased interest in interpretable AI has given rise to a variety of approaches such as feature visualization \cite{matth2013visual}, activation map visualization \cite{WANG2024105646,wang2024enhance} and example-based methods \cite{thia2022training,singla2023explaining}. These methods strive to align AI models with human cognitive processes, thereby enhancing clinicians' trust and facilitating broader adoption into clinical workflows.

The comprehensive review \cite{champendal2023scoping} pointed out that Class Activation Mapping (CAM) visualizations \cite{zhou2016cam} are the most commonly used approach to explain decisions in medical diagnostics. Nevertheless, these visualizations represent a form of post-hoc interpretation. Over-reliance on such post-hoc explanations for high-stakes decisions, particularly in healthcare, may lead to substantial risk \cite{rudin2019stop}. These explanations may produce misleading results as they may not accurately reflect the model's actual decision-making process. This issue is particularly concerning in the medical field where clinicians might overtrust the model's predictions based on seemingly intuitive explanations \cite{Ghassemi2021}.

In this work, we introduce \textit{\cd}, a simple yet effective interpretable diagnosis framework. It leverages contrastive learning approach to implement a case-based reasoning (CBR) mechanism. Explanations by examples have been shown to be the preferred explanation styles according to the non-technical end-users \cite{neurips2020explain}. Our objective is to incorporate CBR mechanism to introduce transparency into diagnostic process. Furthermore, combining with further post-hoc explanation approaches, \cd offers more powerful interpretability for clinicians in medical decisions-making process. 

\paragraph{Related works} In medical diagnosis, CBR system has been instrumental, supporting clinicians by providing insights from historical patient data. These early CBR systems were largely rule-based, capturing the expertise of human specialists to guide the diagnostic process. Pakhomov \etal provided insights into the identification of patient records using institutional databases with example-based and traditional ML techniques \cite{pakhomov2006auto}. More recently, example-based interpretation methods have evolved from CBR's principles, focusing on providing specific instances to explain AI decisions. Chen \etal proposed ProtoPNet, a deep neural network mimics human reasoning by finding similar visual prototypes for classification and demonstrating its efficacy in mammogram analysis \cite{barnett2021interpretable}. Another interpretative strategy involves the use of counterfactual examples to facilitate understanding. Singla \etal proposed a novel method to explain by counterfacts, incorporating generative model to synthesize smoothly changed examples. 

\section{Methods}
\begin{figure}
    \centering
    \includegraphics{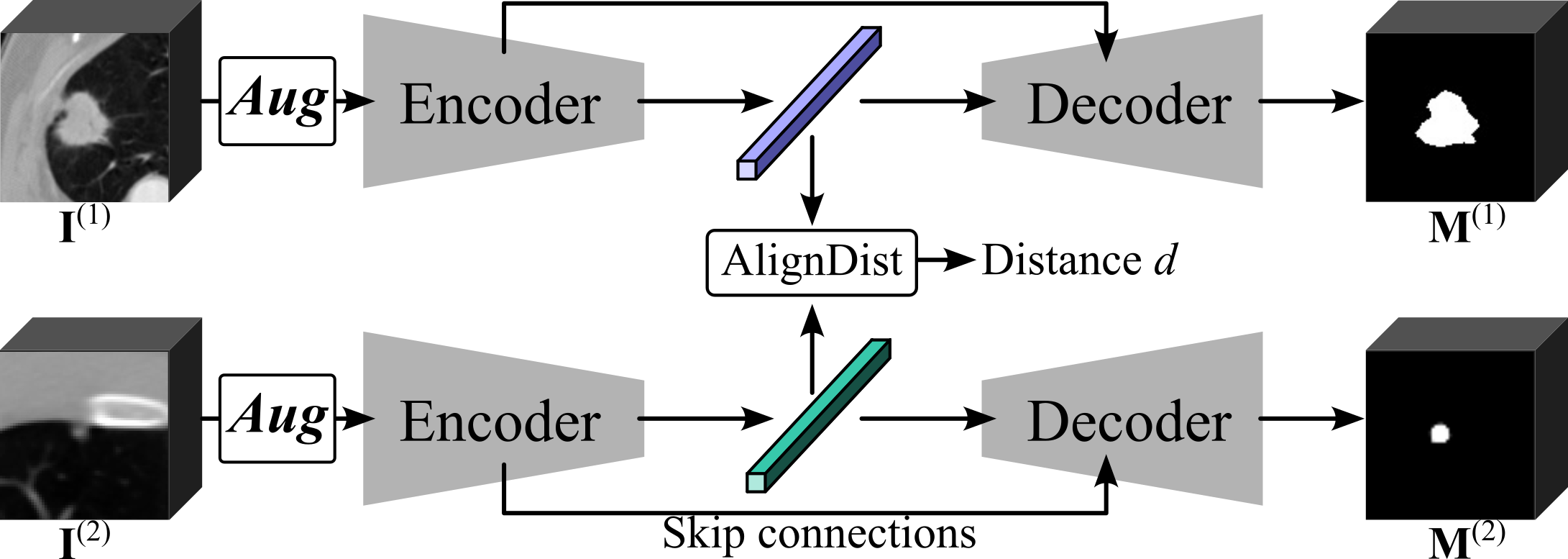}
    \caption{The overview of \cd network architecture. \textit{Aug} denotes online data augmentation. `AlignDist' module is to align two latent representation code and meaure their distance.}
    \label{fig:archi}
\end{figure}

The motivation of this study is to design a transparent and accurate interpretable diagnostic framework that indeed help clinicians to better understand the decision-making logic in `black-box' AI model. The architecture of the proposed \cd is illustrated in Fig.~\ref{fig:archi}.

\subsection{Training}
In this work, we incorporated Siamese network structure as the backbone in the contrastive learning framework, where subnetwork in Siamese structure is a U-Net like neural network. The U-Net like network is designed to segment the lung nodule to introduce an additional attention. The U-Net architecture $S$ can be defined as follows:
\begin{equation}
    S(\mathbf{I}; \theta) := \mathcal{D}\left(\mathcal{E}(\mathcal{A}(\mathbf{I}); \theta_e); \theta_d\right),
\end{equation}
where $\mathcal{E}$ and $\mathcal{D}$ denote the encoder and decoder of the U-Net, with $\theta_e$ and $\theta_d$ representing the weights for the encoder and decoder, respectively. $\mathbf{I}$ is the input images, where $\mathbf{I} \in \mathbb{R}^{H \times W \times C}$. $\mathcal{A}$ denotes online data augmentation.

The loss function for training the \cd consists of two terms: contrastive loss and segmentation loss:
\begin{equation}
    L = \ell_c(\mathbf{c}^{(1)}, \mathbf{c}^{(2)}) + \omega\sum_{k\in(1,2)} \ell_s(\mathbf{\tilde{M}}^{(k)}, \mathbf{M}^{(k)})
\end{equation}
where $\mathbf{c}^{(k)}\in\mathbb{R}^n$ is the  the \textit{n}-dimensional latent representation code from encoder of each subnetwork, which could be obtained from $\mathbf{c}^{(k)} = \mathcal{E}(\mathcal{A}(\mathbf{I}^{(k)})), k\in(1,2)$. $\mathbf{\tilde{M}}^{(k)}, \mathbf{M}^{(k)}$ denote predicted and ground-truth segmentation mask, respectively. $\omega$ is a weight hyper-parameter. $\ell_c$ and $\ell_s$ denote contrastive loss and segmentation loss. DiceCEloss is used as segmentation loss in this work. The contrastive loss is defined as:
\( \ell_c = y\cdot d + (1 - y)\cdot \max(0, m - d) \), $y$ is the label whether two input image belong the same benign/malignant class, $m$ is the margin hyper parameter. The distance $d$ can be obtained by the `AlignDist' module which is defined as:
\begin{equation}
    d = D(\mathbf{A}\mathbf{c}^{(1)} + \mathbf{b}, \mathbf{A}\mathbf{c}^{(2)} + \mathbf{b}),
\end{equation}
where, $D$ is distance metric. $\mathbf{A}\in\mathbb{R}^n$ and $\mathbf{b}\in\mathbb{R}^n$ denote the weight and bias of linear transformation, respectively. The AlignDist is designed to adjust the latent representation that the distance measurement can be made more effectively.

\subsection{Inference}
During the inference, we involve case-based reasoning mechanism to classify the instances. Specifically, we incorporate the \textit{k}-Nearest Neighbors (\textit{k}-NN) algorithm as follows:

Given a query instance \( \mathbf{I}_{\text{qry}} \), we compute its distance to all instances in the support dataset, \ie the labeled training dataset. The \textit{k}-NN algorithm then identifies the \( k \) instances that are closest to \( \mathbf{I}_{\text{qry}} \) based on the chosen distance metric:
\begin{equation}
    \text{NN}(\mathbf{I}_{\text{qry}}) = \underset{\mathbf{I}_i \in \mathrm{\Phi_s}}{\mathrm{argmin}} \, D(\mathbf{I}_{\text{qry}}, \mathbf{I}_i)    
\end{equation}
where $\mathrm{\Phi_s}$ denotes labeled support dataset including both training and validation sets. The labels of these \( k \) nearest neighbors are then aggregated by majority voting, to predict the class label for \( \mathbf{I}_{\text{qry}} \). The hyper-parameter $k$ is optimized on validation set.

\subsection{Explainability}\label{sec:explain}
\cd is grounded in case-based reasoning, which inherently provides the model with significant transparency. This allows clinicians to intuitively understand the decision-making logic of the model by examining comparable cases. Moreover, we also incorporated additional post-hoc interpretation methods to further aid doctors in decision-making. 

\paragraph{Contrastive Visualization} CAM family approaches are the most commonly used for post-hoc explanation visualization. We adopted Local-CAM \cite{wang2024enhance} to pinpoint the paired activation regions across query and support images. Different from conventional Grad-CAM, Local-CAM is specifically designed to highlight more fine-grain activation areas, which aligns with our focused research of lung nodules. We integrated Local-CAM with the Siamese architecture, namely Siamese Local-CAM. It is designed to visualize the differentiated regions of the Siamese network. Gradients are back-propagated from the distance $d$. The computation process can be described as:

\begin{equation}
    L_t(x,y,z)=ReLU\left(\sum_{k}{w_{t-1}(x,y,z)\cdot w_k(x,y,z){\cdot A}_k(x,y,z)}\right),
\end{equation}
where $w_k$ is defined as:
\begin{equation}
    w_k\left(x,y,z\right)=ReLU\left(\frac{\partial d}{\partial A_k(i,j,d)}\left(x,y,z\right)\right),
\end{equation}
where $L_t(x,y)$ is the attention map of target layer $t$. $w_{t-1}(x,y)$ is the weight tensor of upper-level target layer $t-1$. $A_k(x,y,z)$ denotes the activation map of $k$-th channel, $(x,y,z)$ denotes 3D spatial positions. $d$ denotes distance of Siamese network. $A_k(i,j,d)$ denotes voxel located at $(i,j,d)$ position of $A_k$.  

\paragraph{Confidence Score} In addition to visualization, our framework enhances decision-making with a confidence score derived from measuring distances. We set a similarity threshold based on the 95\% confidence interval of distances among correctly matched pairs in the support set. This threshold helps determine the reliability of the cases similar to the query.

\begin{table}[t]
    \centering
    \caption{Comparison results of different lung nodule diagnosis models on LIDC dataset. Quantitative analysis was evaluated by using the area under the ROC curve (\textit{AUC}), accuracy (\textit{Acc}), recall (\textit{Re}), Precision (\textit{Pr}) and F1 score (\textit{F1}).}
    \begin{tabularx}{0.9\textwidth}{X|p{4.5em}p{4.5em}p{4.5em}p{4.5em}p{4.5em}}
    \toprule
        \textbf{Models} & \textit{AUC} & \textit{Acc (\%)} & \textit{Re (\%)} & \textit{Pr (\%)} & \textit{F1 (\%)} \\ \midrule
        ResNet & 0.974 & 94.3 & 87.2 & 79.1 & 82.9 \\ 
        DenseNet & 0.960 & 85.4 & 94.9 & 52.1 & 67.3 \\ 
        HSCNN \cite{SHEN201984} & 0.856 & 84.2 & 70.5 & - & - \\ 
        MC-CNN \cite{shen2017multi} & 0.930 & 87.1 & 77.0 & - & - \\ 
        MSCS-DeepLN \cite{xu2020mscs} & 0.940 & 92.7 & 85.6 & 90.4 & - \\ 
        Fu et al. \cite{fu2022attention} & 0.959 & 94.7 & 96.2 & \textbf{97.8} & - \\ 
        MTMR-Net \cite{liu2019multi} & 0.979 & 93.5 & 93.0 & - & - \\ 
        ExPN-Net \cite{WANG2024105646} & \textbf{0.992} & 95.5 & \textbf{100.0} & 78.0 & 87.6 \\ \hline
        \cd & 0.977 & \textbf{98.0} & 97.4 & 90.5 & \textbf{93.8} \\ \bottomrule
    \end{tabularx}
    \label{result}
\end{table}

\section{Experiments and Results}
In this study, we evaluate \cd using the LIDC/IDRI dataset \cite{armato2011lung}, which is a publicly dataset collected retrospectively from seven academic institutions. The criteria for inclusion and exclusion of data are consistent with those used in related work \cite{WANG2024105646}. Our analysis involves 1,226 lung nodules, which have been randomly divided into two subsets for training (980 nodules) and independent testing (246 nodules). Nodules with a manual assessment score above 3 are classified as malignant, while those with a score of 3 or below are considered benign. 

The preprocessing of the LIDC data starts with loading the 3D images and cropping a local patch with size of [64,64,32] containing the lung nodule. Several data augmentation are randomly applied, including 90-degree rotations, horizontal and vertical flipping, contrast adjustments, affine transformations, and the addition of Gaussian noise. This comprehensive preprocessing routine helps to standardize and diversify the training data, contributing to the effectiveness and generalization power of the subsequent model training.

For the model training, we used L2 distance metric as distance metric. The loss weight $\omega$ for the training is set to 1.5. The optimization algorithm is Adam, initialized with a learning rate of 0.001. Stochastic Gradient Descent with Warm Restarts (SGDR) learning rate scheduling is used. Probability of applying Data augmentation  is 80\%.

\begin{figure}[th]
  \centering
  \includegraphics[width=0.8\textwidth]{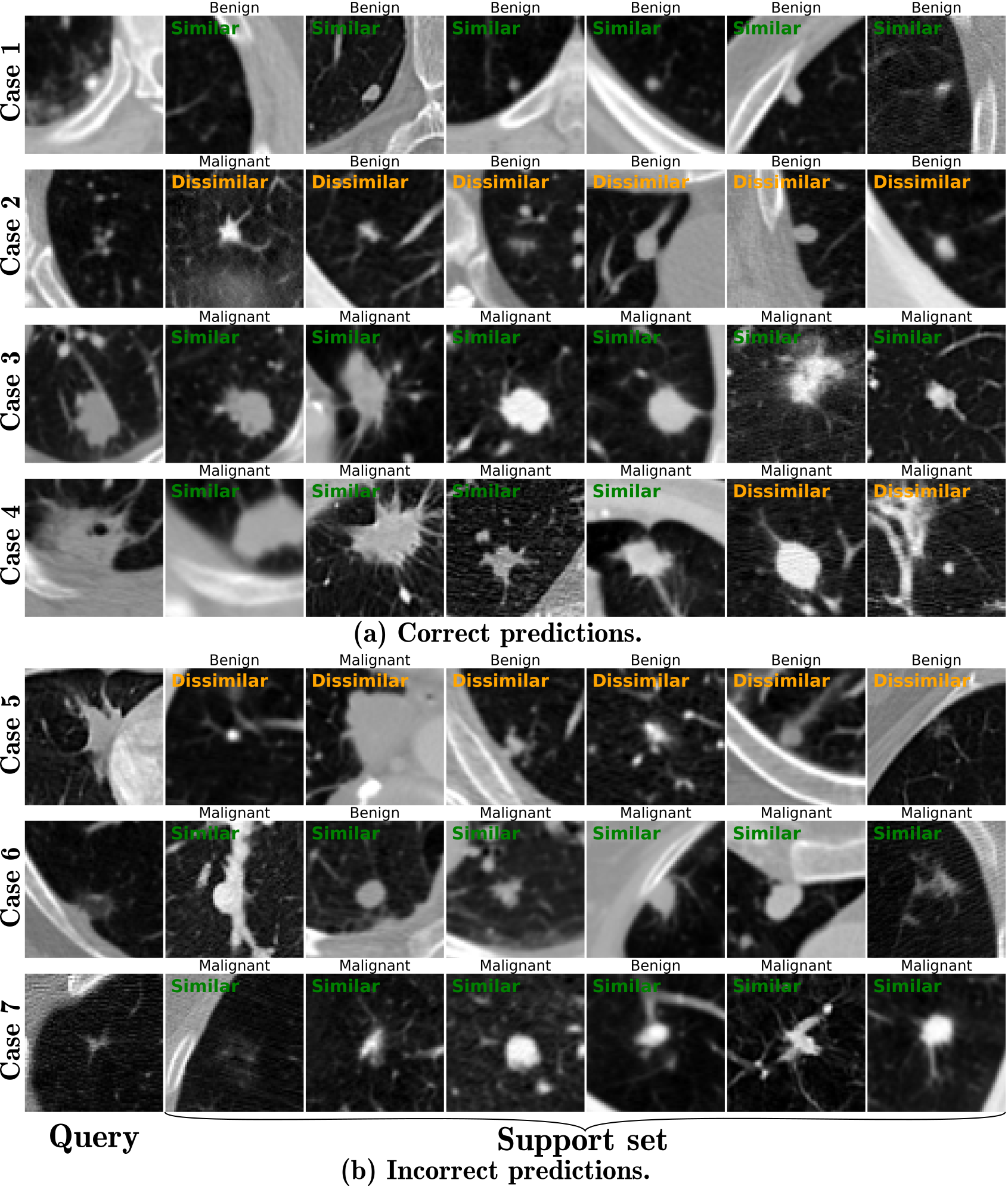} 
  \caption{Several prediction results. The left column shows the query data, while the right-side columns represent the top six most similar cases from support set. The similarity indicators provide a measure of confidence for each prediction.}\label{fig:pairs}
\end{figure}

For diagnostic accuracy comparison, we list several related traditional supervised lung nodule classification works for reference. Quantitive analysis of the results is presented in Table~\ref{result}, which clearly indicates that \cd achieved competitive performance with the related works. Figure~\ref{fig:pairs} illustrates examples of predicted query-support data pairs for both correct and incorrect predictions. The images are marked with similarity indicator reflecting the confidence for each prediction, with the associated method outlined in Section~\ref{sec:explain}.

\begin{figure}[t]
  \newcommand{\w}{0.8\textwidth}
  \centering
  \includegraphics[width=\w]{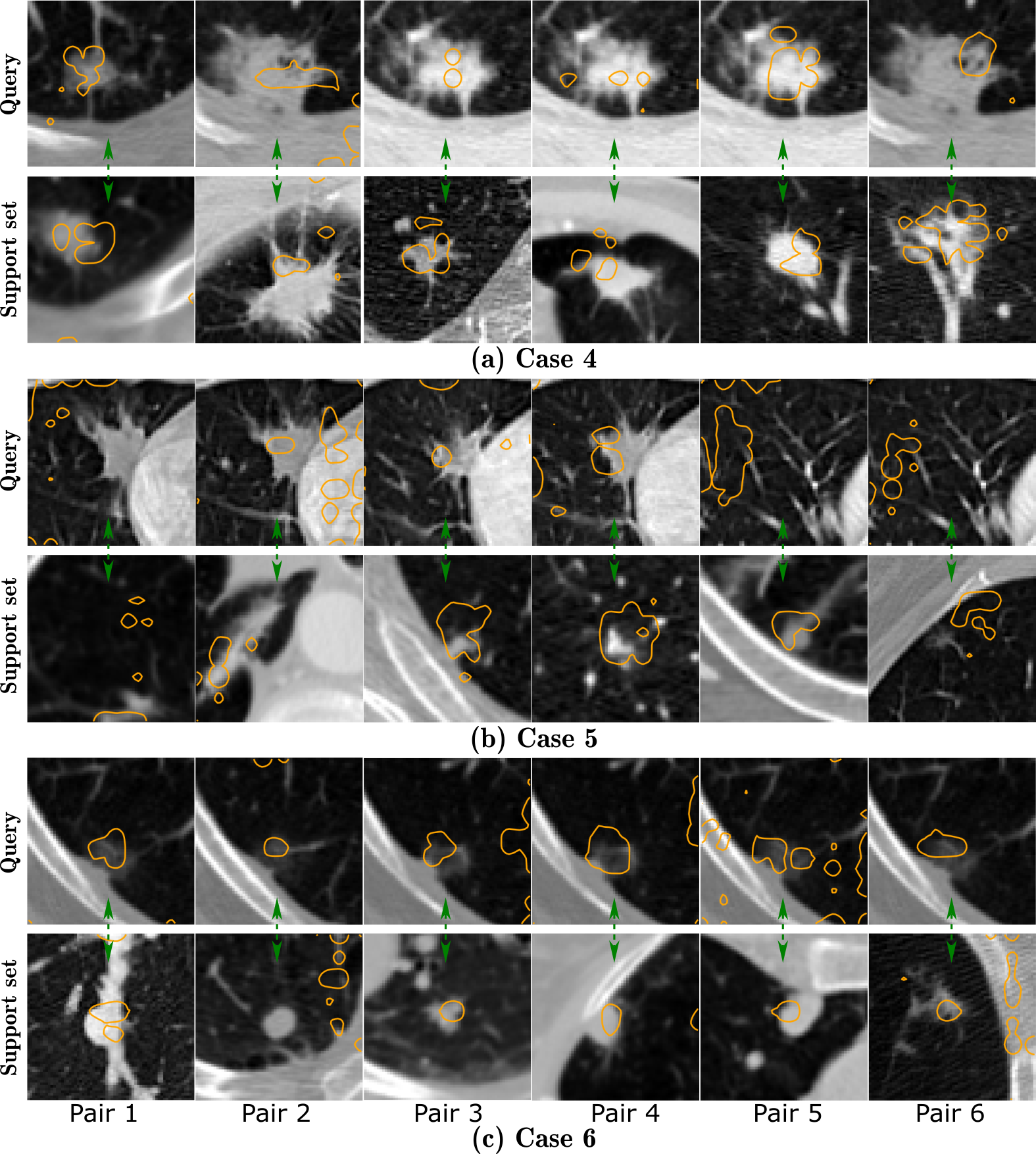}
  \caption{Contrastive regions in query-support pairs. Three cases, aligned with Fig.~\ref{fig:pairs}, are shown. Each column presents a query-support pair with the best slice of 3D patch highlighted, which may appear slightly different from each pair. The region marked with yellow line is the paired activation region identified by Local-CAM.}\label{fig:cam}
\end{figure}

Figure~\ref{fig:cam} illustrates the similar area between corresponding query-support pairs. Three examples are presented, which correspond to the cases in Fig.~\ref{fig:pairs}. To have a best view of similar area, the most appropriate slice of 3D image patch is selected to display, which may appear slightly different from the one in Fig.~\ref{fig:pairs}. Each column indicates a query-support pair as identified by the \cd model, with the region outlined in yellow representing the similar region computed by Local-CAM. More cases are visualized in \textbf{Supplementary}.

\section{Discussions and Conclusion}
It is remarkable to note that \cd obtained top-tier level of diagnosis accuracy compared to other supervised classification approaches. The F1 score demonstrates that \cd have superior balanced performance compared to other methods. A notable constraint of the CBR technique lies in its limited variation within the support set. This shortcoming is depicted in Case 2 illustrated in Fig.~\ref{fig:pairs}, where despite correctly classified, dissimilar cases are selected. We believe that the high accuracy is primarily due to the leverage of contrastive learning methods, which contribute to a smoother and more consistent latent space.

By providing similar cases, clinicians can intuitively understand the decision-making logic of the model. To further aid clinicians' diagnosis, we provide a similarity indicator derived from the distribution of distances within the support set, yielding a measure of confidence when assessing query data. The introduction of Siamese Local-CAM further aids this process by providing visual representations of regions deemed similar for each query-support pair. This supplementary information is designed to enable efficient verification of the similarity regions proposed by the model. In practice, such as demonstrated by Case 5, clinicians are enable to effectively challenge the model's conclusion by examining the overlaid similarity region and its associated confidence score.

To address the constraints posed by the limited variation within the support set, our future work will explore the integration of generative models capable of synthesizing smooth representations of similar cases. Generative model is designed to generate a broader and more detailed spectrum of case-specific data that would act as a comparative set for enhancing the model's diagnostic precision. To simplify the visual presentation would be another future work. Currently, the visual presentation is a little bit complex. Simpler presentation with fewer visual elements could offer clinicians a clearer, more intuitive understanding. Moreover, leveraging large datasets, such as the NLST, will be pivotal in overcoming current limitations.

In conclusion, we propose \cd framework, a straightforward yet powerful method for interpretable diagnostics. It utilizes contrastive learning methodologies to establish a case-based reasoning mechanism, which naturally forms a transparent model. Moreover, \cd enhances interpretability with post-hoc explanations such as activation-map visualization and confidence scoring to further aid clinicians to understand the inherent decision-making of the AI model.

%
%
\bibliographystyle{splncs04}
\bibliography{ref_short}

\end{document}